\documentclass[twocolumn,english,conference]{IEEEtran}
\usepackage[T1]{fontenc}
\usepackage[latin9]{inputenc}
\usepackage{amsmath}
\usepackage{amssymb}
\usepackage{graphicx}

\makeatletter

\providecommand{\tabularnewline}{\\}

\usepackage{cite}

\usepackage{babel}
\usepackage{subfig}
\usepackage{caption}

\makeatother

\usepackage{babel}
\begin{document}

\title{Real-Time Road Segmentation Using LiDAR Data Processing on an FPGA}

\author{Yecheng Lyu, Lin Bai, and Xinming Huang\\
 Department of Electrical and Computer Engineering\\
Worcester Polytechnic Institute\\
Worcester, MA 01609, USA\\
\{ylyu,lbai2,xhuang\}@wpi.edu}
\maketitle
\begin{abstract}
This paper presents the FPGA design of a convolutional neural network
(CNN) based road segmentation algorithm for real-time processing of
LiDAR data. For autonomous vehicles, it is important to perform road
segmentation and obstacle detection such that the drivable region
can be identified for path planning. Traditional road segmentation
algorithms are mainly based on image data from cameras, which is subjected
to the light condition as well as the quality of road markings. LiDAR
sensor can obtain the 3D geometry information of the vehicle surroundings
with very high accuracy. However, it is a computational challenge
to process a large amount of LiDAR data at real-time. In this work,
a convolutional neural network model is proposed and trained to perform
semantic segmentation using the LiDAR sensor data. Furthermore, an
efficient hardware design is implemented on the FPGA that can process
each LiDAR scan in 16.9ms, which is much faster than the previous
works. Evaluated using KITTI road benchmarks, the proposed solution
achieves high accuracy of road segmentation. \end{abstract}


\section{Introduction}

In recent years, we have witnessed a strong increase of research interests
on advanced driver assistance systems (ADAS) and autonomous vehicles.
While fully autonomous driving might still be years away, there are
many recent research on traffic scene perception and its implementations
on various platforms. The traffic scene perception task can be separated
into two sub-tasks: object detection and road/lane detection. Object
detection includes vehicle detection\cite{xiang2017subcategory}\cite{Forward_Collision_Warning}\cite{ren2015faster}\cite{gonzalez2015multiview}\cite{teichmann2016multinet},
pedestrian detection\cite{Pedestrian_Detection}\cite{ren2015faster}\cite{gonzalez2015multiview}
and traffic light/sign detection\cite{chen2016accurate}\cite{chen2014gpu}\cite{zhao2014efficient}\cite{rajaram2016refinenet}\cite{Traffic_Sign_Detection},
while road/lane detection includes road marking detection\cite{chen2015road}\cite{Road_Marking_Detection}\cite{aly2008real},
lane detection\cite{zhao2014real}\cite{beyeler2014vision}\cite{chen2017end}
and road segmentation\cite{caltagirone2017fast}\cite{laddha2016map}\cite{beyeler2014vision}.
In this work, we are primarily concentrating on road segmentation,
since it is a fundamental component of automated driving that provides
the drivable region for the vehicle's next movement. 

Many sensing modalities have been used for road segmentation. Monocular
vision\cite{beyeler2014vision}\cite{teichmann2016multinet}\cite{oliveira2017efficient}
and stereo vision\cite{vitor2014probabilistic}\cite{einecke2014block}
are mostly used because cameras are low-cost and have a similar view
to human eyes. However, considering road appearance diversity, image
clarity issues, poor visibility conditions\cite{Road_Lane_Detection_Survey1},
image-based feature describers are often difficult to generate and
easy to fail. In contrast of passive sensors such as cameras, light
detection and ranging (LiDAR) actively emits laser beams and measures
the distance from the reflection by time of flight (TOF). Therefore,
LiDAR is robust to environmental illumination. Several recent works
have studied road segmentation based on LiDAR information or the combination
of LiDAR and camera data\cite{xiao2015crf,xiao2017hybrid}. 

For the applications of autonomous vehicles, both real-time performance
and power consumption need to be considered\cite{ADAS_Survey}. Graphic
Processing Unit (GPU) is a popular platform for parallel processing,
but power consumption is usually high. FPGA suits to the condition
with limited power supply, such as an autonomous vehicle. Moreover,
FPGA can be developed as a customized integrated circuit that can
perform massive parallel processing and data communications on-chip.
Hereby, we propose to target the LiDAR based road segmentation algorithm
on an FPGA as a real-time low-power embedded system. 

In this paper, the problem of road segmentation is framed as a semantic
segmentation task in spherical image using a deep neural network.
Instead of an encoder-decoder structure often implemented in traditional
neural networks, a block containing a convolutional layer and a non-linear
layer is cascaded twelve times so that multiplexing can be applied
on the processing blocks on-chip. The proposed solution is evaluated
on KITTI benchmarks and achieve satisfactory result. The rest of paper
is organized as follows. Section \ref{sec:Related-Work} introduces
the related work of road perception problem. The proposed convolutional
neural network (CNN) structure and its performance on KITTI benchmarks
are presented in Section \ref{sec:Algorithms-Design}. Section \ref{sec:Hardware-Architecture}
presents the FPGA design hardware architecture and implementation
results. Finally Section \ref{sec:Conclusion} concludes the paper.

\section{Related Work\label{sec:Related-Work}}

Road segmentation has been studied with different sensors and algorithms
over the past decade. In the early years, researchers used manually
designed feature descriptors to separate the road from others. At
that time, camera was the major sensor and features were often generated
based on the illumination and shape from images\cite{aly2008real}\cite{beyeler2014vision}\cite{kong2010general},
which led to low accuracy and the performance variations from different
light conditions and road scenes. Recently, two major techniques have
been investigated to overcome the shortcomings of manually selected
features in images. One is to use machine learning to design a complex
and robust feature descriptor, such as CNN\cite{mendes2016exploiting}\cite{gonzalez2015multiview}\cite{oliveira2017efficient}
and conditional random field (CRF)\cite{passani2014crf}. The other
is to use intensity invariant sensor or multi-sensor fusion instead
of camera to obtain a more robust descriptor. A popular intensity
invariant sensor is LiDAR\cite{huang2009finding}. There were also
research works trying to combine those two and apply machine learning
to data processing. Several results showed high accuracy, but their
processing time is too long to be employed for real-time applications\cite{xiao2017hybrid}\cite{xiao2015crf}.
For autonomous driving, road segmentation must be implemented on real-time
embedded platforms such as FPGA, application-specific integrated circuit
(ASIC), or a mobile CPU/GPU processor.  A neural network was proposed
in\cite{huval2015empirical} to detect lane markers on the road and
had the run-time of 2.5Hz on TK1 mobile GPU platform. Similarly, research
work in\cite{romera2017efficient} proposed a neural network to segment
multiple objects including vehicle, pedestrian and pavement and achieved
10Hz run time at the resolution of 480-by-320 pixels on TX1 GPU platform.
In\cite{zhao2014real} and\cite{wang2013fpga}, FPGA based solutions
are proposed for lane detection and resulted 60Hz and 550Hz processing
speed, respectively.

\section{Algorithms Design\label{sec:Algorithms-Design}}

The goal of road segmentation is to label the drivable region, also
called free space. The input data comes from different sensors such
as camera, LiDAR, GPS and IMU. The output are usually presented as
area on the top-view or labeled pixels on camera view. In this paper,
we choose LiDAR data as input, a deep neural network as the processor,
and top-view predictions as the main output to evaluate the road segmentation
performance. Results on camera view are also presented for better
visualization. The proposed algorithm has the following three steps:
pre-processing, neural network processing and post-processing.

\subsection{Pre-processing}

During pre-processing, input data points are arranged and projected
into a 3-D blob with \emph{M} by \emph{N} tensors and \emph{C} channels
so that the tensor can flow through the layers in the neural network
to produce an output. A tensor refers to a specific view in the real
world. There are four types of views available for the autonomous
driving task: image view (also known as camera view), top view (also
known as bird eye view), cylindrical view and spherical view. Image
view and top view are commonly choices, because in this two views
LiDAR data can be fused with camera data and those views are natural
to human eyes. However, LiDAR points are sparse in those views. Statistically,
LiDAR points covers only 4\% of pixels in image view and 5.6\% on
top-view. That means majority inputs into the neural network are zero
input leading to waste of computing resource. Cylindrical view and
Spherical view match the LiDAR sensing scheme and data points can
cover up to 91\% pixels on the map. Hereby we choose spherical view
as the projection scheme. The resolution of polar angle $\theta$
and azimuthal angle $\varphi$ are chosen based on the LiDAR resolution.
In this work, all 64 rows are included in vertical. While in horizontal
LiDAR points are grouped by 0.4$\textdegree$ which doubles the designed
resolution of LiDAR to minimum the number of cells without LiDAR measurements.
The input blob has 256 columns and FOV is shifting to augment training
data.

\begin{figure}
\begin{centering}
\includegraphics[width=0.95\columnwidth]{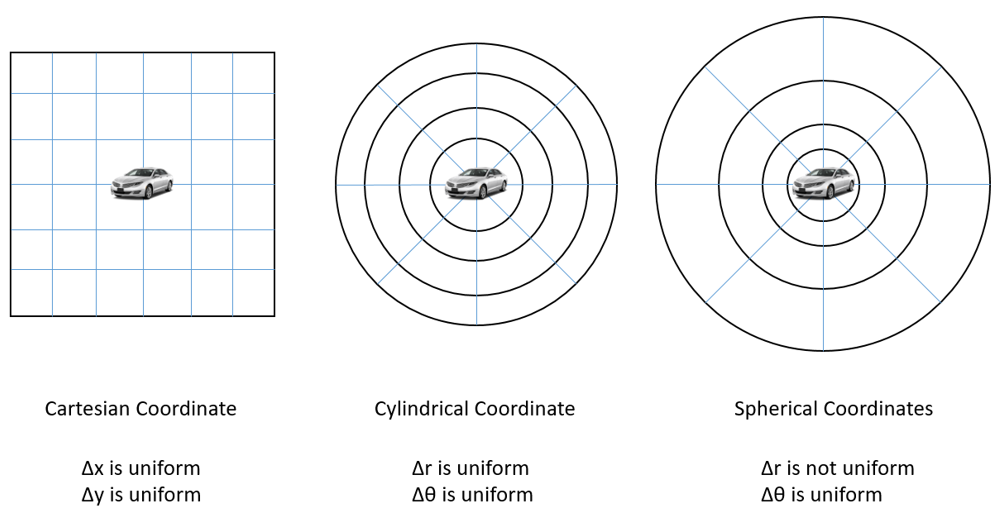}
\par\end{centering}

\caption{Grid projection to ground from different views}

\end{figure}

Although spherical view is chosen for data projection, we can still
add additional feature channels from other views to improve the accuracy
of the trained neural network. Here we select sixteen channels, the
first 7 channels come from the LiDAR point which has the lowest altitude
in the cell, the next 7 channels come from the LiDAR point which has
the highest altitude. The 7 channels are location of measured points
in Cartesian coordinate (\emph{x},\emph{y},\emph{z}), location of
measured points in spherical coordinate ($\theta,\varphi,r$), and
reflection intensity of measured points (\emph{H}). The other 2 channels
are the location of cell on the 2D map (\emph{i}, \emph{j}).

\begin{figure}
\begin{tabular}{cc}
\includegraphics[width=0.45\columnwidth]{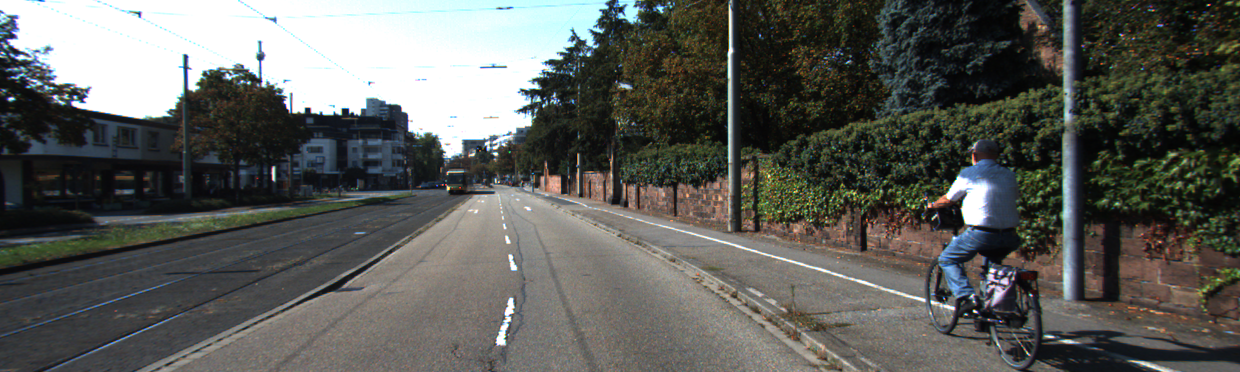} & \includegraphics[width=0.45\columnwidth]{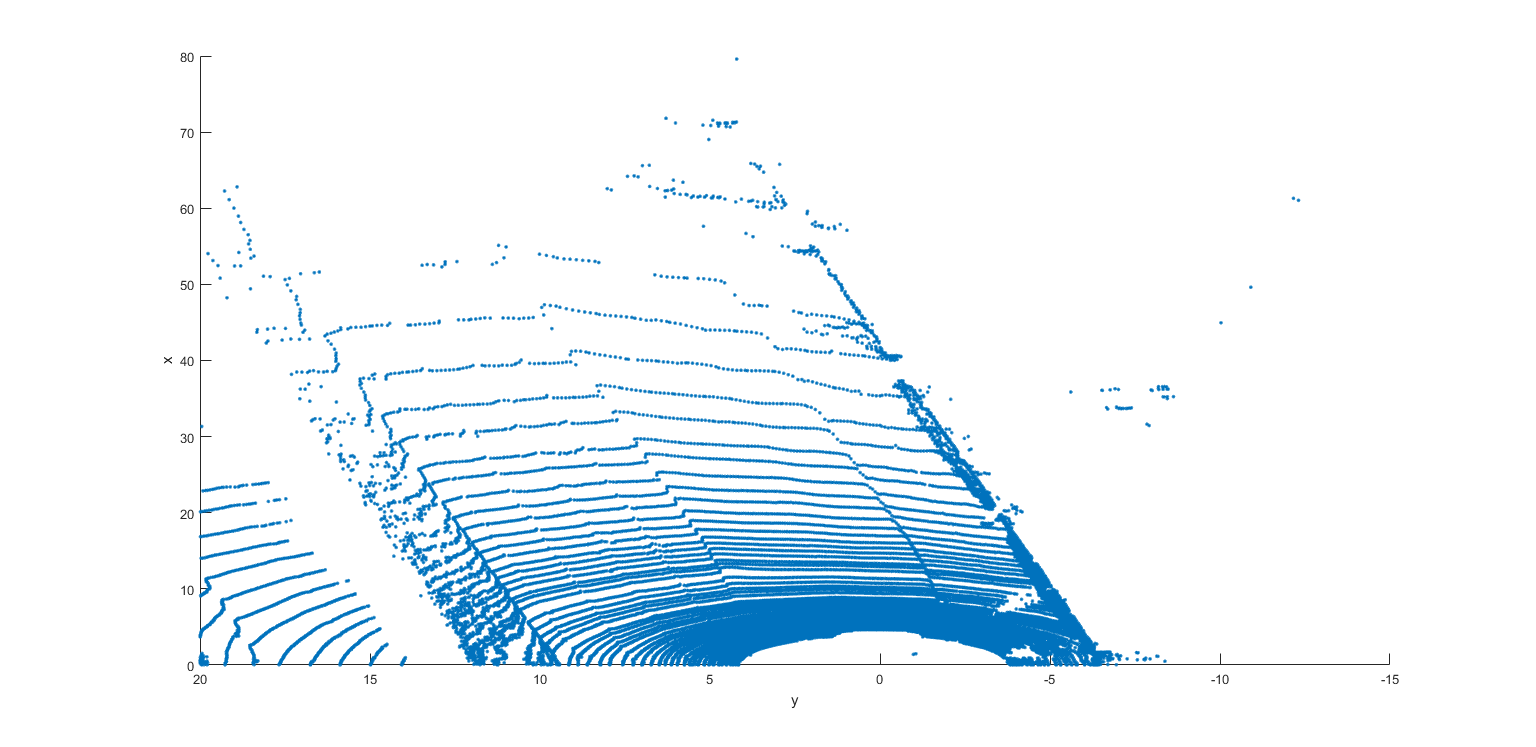}\tabularnewline
\end{tabular}

\caption{An illustration of camera view and the corresponding LiDAR points.}
\end{figure}

\begin{figure}
\begin{centering}
\includegraphics[width=0.95\columnwidth]{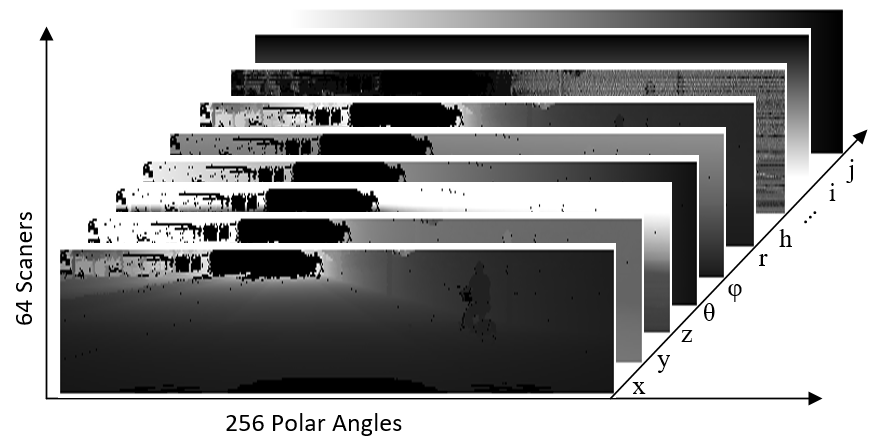}
\par\end{centering}

\caption{Input map to the neural network with 9 channels.}
\end{figure}

\subsection{Neural network processing}

\begin{figure}
\begin{centering}
\includegraphics[width=0.95\columnwidth]{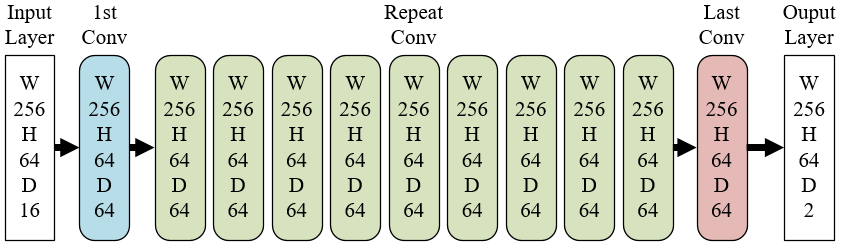}
\par\end{centering}

\caption{\label{fig:Architecture-of-Nerual}Architecture of the convolutional
neural network for road segmentation.}
\end{figure}

In autonomous driving, traffic scene perception is often implemented
on embedded systems. In consideration of limited computational resource
in an embedded system, we proposed a new network architecture that
minimize the GPU and FPGA memory by multiplexing the blob memory.
The architecture is shown in Figure \ref{fig:Architecture-of-Nerual}.
Except for the first and last convolutional layers, those 9 layers
in between are constructed using the same structure. Each repetitive
structure includes a convolutional layer and an activation layer.
The convolutional layer is built with 64 filters and each filter has
a $5\times5$ kernel with stride size of 1 and padding size of 2.
The stride and padding settings make output of the convolutional has
the same size as the input. Rectified linear unit is chosen as the
activation function for fast training. Two drop-out layers are added
after 6th block and 10th block in training phase to accelerate convergence.
It can be seen that there is no pooling layers and all blobs have
the same size except for the input blob and score blob. Therefore
all internal results can be stored in the same memory space directory
without allocation or reshaping the blob. We choose $5\times5$ convolutional
kernel size and 11 convolutional layers from our experimental results
that this settings is a good trade-off between CNN performance and
resource usage.

\subsection{Post-processing}

\begin{figure}
\begin{tabular}{cc}
\includegraphics[width=0.6\columnwidth]{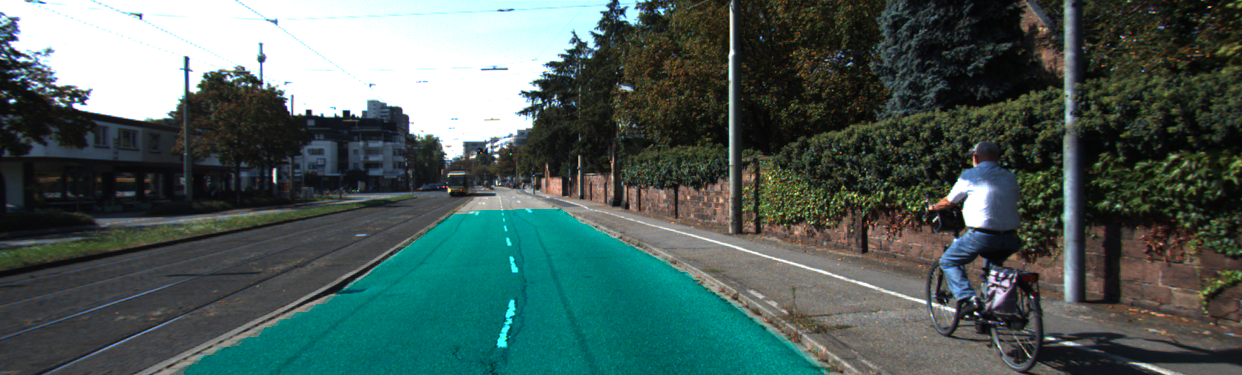} & \includegraphics[width=0.35\columnwidth]{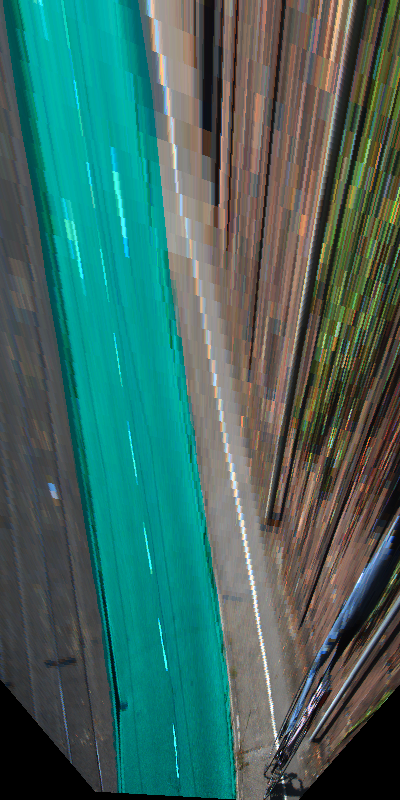}\tabularnewline
\end{tabular}

\caption{\label{fig:Drivable-area-on}Drivable area on camera view and top
view projected from the neural network output}
\end{figure}

In post-processing, results obtained from the neural network are projected
back to targeted views, i.e. camera view and top view, for performance
validation. The challenge of the post-processing is that the points
in the output of neural network are non-uniformly distributed on the
target view after projection. Traditional image processing methods,
such as dilation, erosion, closing and opening, are not able to generate
a filled area with smoothed contour. In our post-processing step,
contour of the drivable area is firstly determined and then the region
within the contour is marked as the segmentation results on target
view. Figure \ref{fig:Drivable-area-on} shows an example of the road
segmentation results.

To determine the contour of the drivable area, the furthest points
in each angle $\theta$, which is corresponding to the each column
of the neural network output, are selected and projected onto the
target view. Subsequently, a polyline is drawn along those furthest
points on all angles on the target view. The polyline graph becomes
a polygon if we add a straight line at the bottom. The polygon is
then treated as contour of drivable area and filled up with semantic
pixel labels.

\subsection{Training and evaluation on KITTI road benchmark}

To evaluate the performance of the proposed approach, we train the
network on KITTI road/lane detection dataset. As described in \cite{KITTI_Road_Benchmark},
maximum F1 score ($F_{max}$) and average precision (AP) are the key
measurement to evaluate the performance of road perception algorithms.
$F_{max}$ provides the insight of an algorithm's optimal performance,
while AP indicates its average performance. In Table \ref{tab:Comparison-of-result},
we compared our proposed approach with several results published recently.
It shows that our proposed approach has comparable performance but
uses significantly less processing time. The actual processing time
of the neural network implemented on the FPGA is about 16.9ms. Since
most of the execution times listed in Table 1 were from various GPU
platforms, we also evaluate our algorithm on a K20 GPU using MATLAB
on Caffe and the total processing time is about 120ms, including pre-processing,
neural network, post-processing, and visualization.

\begin{table}
\begin{centering}
\caption{\label{tab:Comparison-of-result}Comparison with existing results
on KITTI road/lane detection dataset.}

\par\end{centering}

\centering{}%
\begin{tabular}{|c|r|c|r|}
\hline 
Name & $F_{max}$ & AP & run time\tabularnewline
\hline 
This work on FPGA & 91.79\% & 84.76\% & 16.9ms\tabularnewline
\hline 
HybridCRF\cite{xiao2017hybrid} & 90.81\% & 84.79\% & 1500ms\tabularnewline
\hline 
LidarHisto\cite{chen2017lidar} & 90.67\% & 84.79\% & 100ms\tabularnewline
\hline 
MixedCRF & 90.59\% & 84.24\% & 6000ms\tabularnewline
\hline 
FusedCRF\cite{xiao2015crf} & 88.25\% & 79.24\% & 2000ms\tabularnewline
\hline 
RES3D-Velo\cite{shinzato2014road} & 86.58\% & 78.34\% & 360ms\tabularnewline
\hline 
\end{tabular}
\end{table}

\section{Hardware Architecture\label{sec:Hardware-Architecture}}

\begin{figure}
\begin{centering}
\includegraphics[width=1\columnwidth]{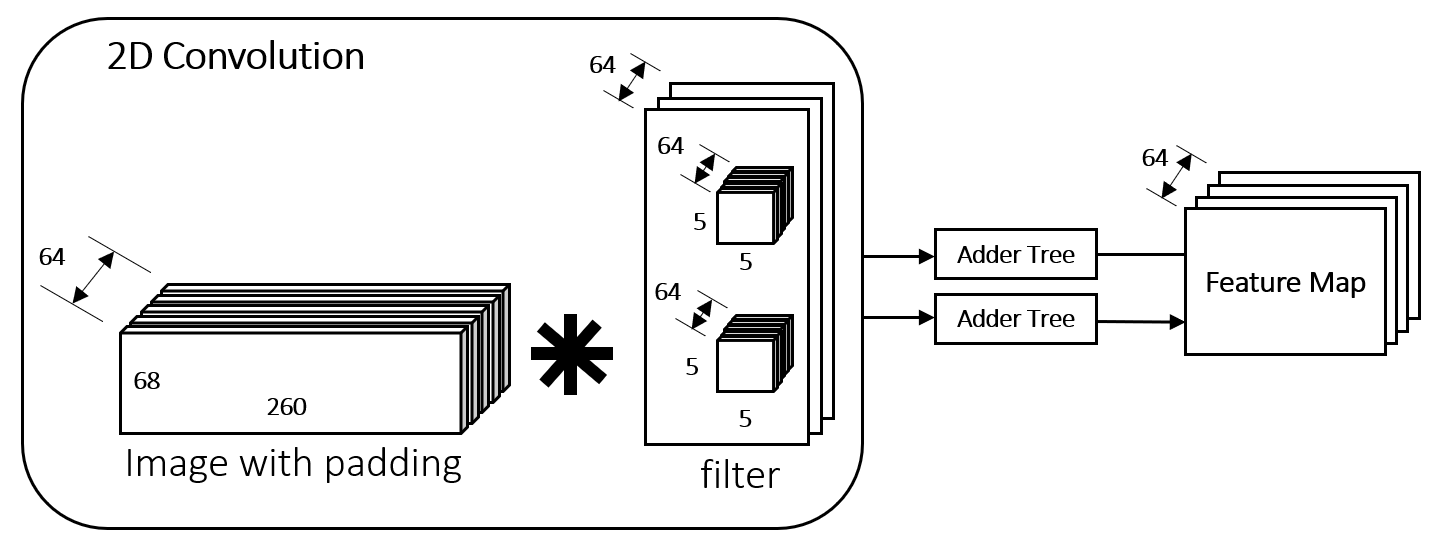}\caption{\label{fig:Hardware-Architecture-Overview}Hardware architecture of
the implementation of convolution layer.}

\par\end{centering}

\end{figure}

As described in Section \ref{sec:Algorithms-Design}, we organize
the LiDAR data into an image map with 16 channels in the size of $256\times64$.
The block diagram of the convolutional layer architecture is shown
in Figure \ref{fig:Hardware-Architecture-Overview}. The same convolutional
unit is used repetitively. There are totally 64 memories to store
intermediate feature map and each memory size is 256k bits. The large
3D convolution can be broken into 64 parallel 2D convolutions, each
with 2 filters, followed by an adder tree to generate the feature
map. 

\subsection{Zero padding generation}

\begin{figure}
\centering{}\includegraphics[width=0.8\columnwidth]{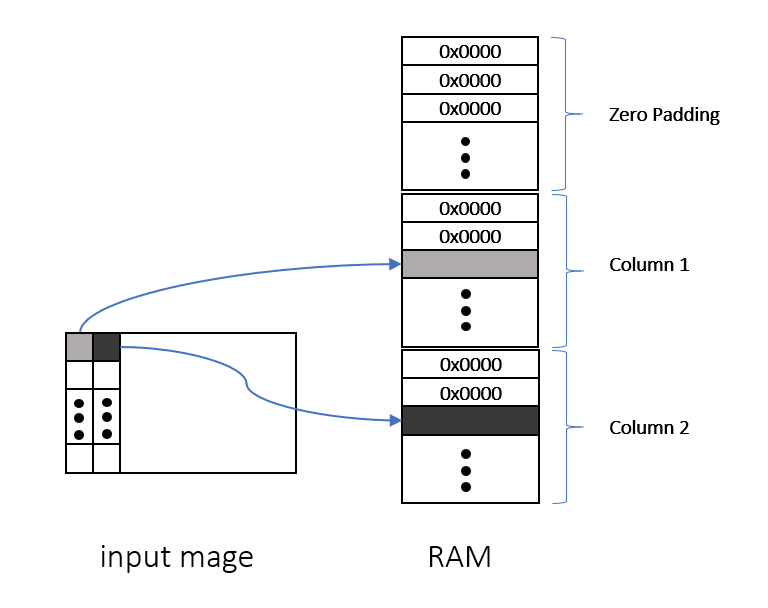}\caption{\label{fig:Zero-Padding-RAM}An illustration of the zero-padding in
the RAM}
\end{figure}

Zero padding helps to control the size of feature maps and to reserve
the boundary information of the input images in convolution operation.
Since the input images are transmitted without padding, a special
dual-port RAM is designed for the convenience of the next stage convolution.
As shown in Figure \ref{fig:Zero-Padding-RAM}, each slot of RAM represents
one column of the input image. The padded zeros are stored in the
block RAM in advance. Control logic is used to store each pixel into
proper memory location. On the other side of the memory, a scanning
circuit reads out data from this RAM pixel by pixel.  

\subsection{2D convolution}

\begin{figure}
\begin{centering}
\includegraphics[scale=0.25]{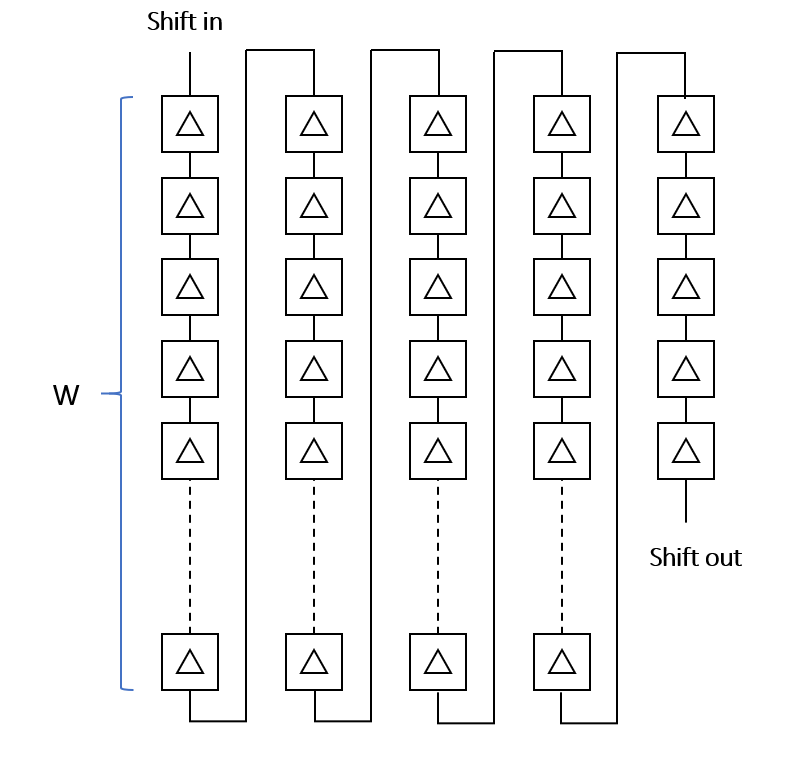}
\par\end{centering}

\caption{\label{fig:Line-Buffer}Line buffer for 2D convolution}
\end{figure}

2D convolution is implemented in conjunction with a line buffer which
consists of 4 lines and 5 additional registers. As demonstrated in
Figure \ref{fig:Line-Buffer}, it outputs $5\times5$ pixel window
in parallel for the multiplication with the weight matrix using 25
multipliers. A highly pipelined adder tree follows the multiplication
to compute the sum. 

\subsection{Control logic}

Because of the large RAM consumption for images with zero padding
and feature maps, a loop-based control is proposed. Each 2D convolution
could generate 2 feature maps. One finite state machine is used to
generate 64 feature maps in 32 loops, reusing the block RAM for images
with padding. To achieve 11 layers of convolution, another finite
state machine is implemented for loop controlling. Therefore, completing
the 11 fully convolutional layers with each depth of 64 requires to
perform the 2D convolutions 352 times.

\subsection{Implementation results \label{sec:result}}

We implement this fully convolutional network on Xilinx UltraScale
XCKU115 FPGA. The targeted operating frequency is set to 350MHz. Each
2D convolution takes about 18,000 clock cycles. It takes about 16.9ms
to complete all 11 convolutional layers, each with filter depth of
64 (except for depth of 16 in the first layer and the depth of 2 in
the last layer). Since LiDAR normally scans at 10Hz, this FPGA implementation
fulfills the requirement of real-time processing. When tested on the
Intel Xeon CPU E5-2687W v3, the processing time is about 500ms. Therefore,
the FPGA implementation gains the speedup factor of 30 over CPU. The
resource usage of the FPGA implementation is listed in Table \ref{tab:Resource-Usage}. 

\begin{table}
\caption{\label{tab:Resource-Usage}Resource usage of the neural network implementation
on an FPGA}

\centering{}%
\begin{tabular}{lccc}
\hline 
 & Used & Available & Utilization\tabularnewline
\hline 
Slice Registers & 43726 & 1326720 & 3.30\%\tabularnewline
\hline 
Slice LUTs & 18684 & 663360 & 2.82\%\tabularnewline
\hline 
Block RAM Tile & 1513 & 2688 & 70.05\%\tabularnewline
\hline 
DSPs & 4480 & 5520 & 81.16\%\tabularnewline
\hline 
\end{tabular}
\end{table}

\section{Conclusions and Future Work \label{sec:Conclusion}}

In this paper, we propose a neural network based approach for road
segmentation using LiDAR data. The neural network is trained with
KITTI road/lane detection dataset and evaluated on its test benchmark.
Moreover, the proposed fully connected neural network is implemented
on an FPGA for real-time low-power processing, which results the processing
time of only 16.9ms for each LiDAR scan. The implementation consumes
a large amount of FPGA on-chip memory. 

For future work, we are considering using the external DDR4 SDRAM
to store feature maps. We also notice during testing that sidewalk
and railway with same altitude as road pavement contributes to the
majority of false positive. Fusion of LiDAR and camera data is needed
to further improve the accuracy.

\bibliographystyle{IEEEtran}
\bibliography{reference}

\begin{thebibliography}{10}
\providecommand{\url}[1]{#1}
\csname url@samestyle\endcsname
\providecommand{\newblock}{\relax}
\providecommand{\bibinfo}[2]{#2}
\providecommand{\BIBentrySTDinterwordspacing}{\spaceskip=0pt\relax}
\providecommand{\BIBentryALTinterwordstretchfactor}{4}
\providecommand{\BIBentryALTinterwordspacing}{\spaceskip=\fontdimen2\font plus
\BIBentryALTinterwordstretchfactor\fontdimen3\font minus
  \fontdimen4\font\relax}
\providecommand{\BIBforeignlanguage}[2]{{%
\expandafter\ifx\csname l@#1\endcsname\relax
\typeout{** WARNING: IEEEtran.bst: No hyphenation pattern has been}%
\typeout{** loaded for the language `#1'. Using the pattern for}%
\typeout{** the default language instead.}%
\else
\language=\csname l@#1\endcsname
\fi
#2}}
\providecommand{\BIBdecl}{\relax}
\BIBdecl

\bibitem{xiang2017subcategory}
Y.~Xiang, W.~Choi, Y.~Lin, and S.~Savarese, ``Subcategory-aware convolutional
  neural networks for object proposals and detection,'' in \emph{Applications
  of Computer Vision (WACV), 2017 IEEE Winter Conference on}.\hskip 1em plus
  0.5em minus 0.4em\relax IEEE, 2017, pp. 924--933.

\bibitem{Forward_Collision_Warning}
E.~Dagan, O.~Mano, G.~P. Stein, and A.~Shashua, ``Forward collision warning
  with a single camera,'' in \emph{Intelligent Vehicles Symposium, 2004
  IEEE}.\hskip 1em plus 0.5em minus 0.4em\relax IEEE, 2004, pp. 37--42.

\bibitem{ren2015faster}
S.~Ren, K.~He, R.~Girshick, and J.~Sun, ``Faster r-cnn: Towards real-time
  object detection with region proposal networks,'' in \emph{Advances in neural
  information processing systems}, 2015, pp. 91--99.

\bibitem{gonzalez2015multiview}
A.~Gonz{\'a}lez, G.~Villalonga, J.~Xu, D.~V{\'a}zquez, J.~Amores, and A.~M.
  L{\'o}pez, ``Multiview random forest of local experts combining rgb and lidar
  data for pedestrian detection,'' in \emph{Intelligent Vehicles Symposium
  (IV), 2015 IEEE}.\hskip 1em plus 0.5em minus 0.4em\relax IEEE, 2015, pp.
  356--361.

\bibitem{teichmann2016multinet}
M.~Teichmann, M.~Weber, M.~Zoellner, R.~Cipolla, and R.~Urtasun, ``Multinet:
  Real-time joint semantic reasoning for autonomous driving,'' \emph{arXiv
  preprint arXiv:1612.07695}, 2016.

\bibitem{Pedestrian_Detection}
D.~M. Gavrila and S.~Munder, ``Multi-cue pedestrian detection and tracking from
  a moving vehicle,'' \emph{International journal of computer vision}, vol.~73,
  no.~1, pp. 41--59, 2007.

\bibitem{chen2016accurate}
Z.~Chen and X.~Huang, ``Accurate and reliable detection of traffic lights using
  multiclass learning and multiobject tracking,'' \emph{IEEE Intelligent
  Transportation Systems Magazine}, vol.~8, no.~4, pp. 28--42, 2016.

\bibitem{chen2014gpu}
Z.~Chen, X.~Huang, Z.~Ni, and H.~He, ``A gpu-based real-time traffic sign
  detection and recognition system,'' in \emph{Computational Intelligence in
  Vehicles and Transportation Systems (CIVTS), 2014 IEEE Symposium on}.\hskip
  1em plus 0.5em minus 0.4em\relax IEEE, 2014, pp. 1--5.

\bibitem{zhao2014efficient}
J.~Zhao, X.~Huang, and Y.~Massoud, ``An efficient real-time fpga implementation
  for object detection,'' in \emph{New Circuits and Systems Conference
  (NEWCAS), 2014 IEEE 12th International}.\hskip 1em plus 0.5em minus
  0.4em\relax IEEE, 2014, pp. 313--316.

\bibitem{rajaram2016refinenet}
R.~N. Rajaram, E.~Ohn-Bar, and M.~M. Trivedi, ``Refinenet: Iterative refinement
  for accurate object localization,'' in \emph{Intelligent Transportation
  Systems (ITSC), 2016 IEEE 19th International Conference on}.\hskip 1em plus
  0.5em minus 0.4em\relax IEEE, 2016, pp. 1528--1533.

\bibitem{Traffic_Sign_Detection}
A.~De~La~Escalera, L.~E. Moreno, M.~A. Salichs, and J.~M. Armingol, ``Road
  traffic sign detection and classification,'' \emph{IEEE transactions on
  industrial electronics}, vol.~44, no.~6, pp. 848--859, 1997.

\bibitem{chen2015road}
T.~Chen, Z.~Chen, Q.~Shi, and X.~Huang, ``Road marking detection and
  classification using machine learning algorithms,'' in \emph{Intelligent
  Vehicles Symposium (IV), 2015 IEEE}.\hskip 1em plus 0.5em minus 0.4em\relax
  IEEE, 2015, pp. 617--621.

\bibitem{Road_Marking_Detection}
T.~Wu and A.~Ranganathan, ``A practical system for road marking detection and
  recognition,'' in \emph{Intelligent Vehicles Symposium (IV), 2012
  IEEE}.\hskip 1em plus 0.5em minus 0.4em\relax IEEE, 2012, pp. 25--30.

\bibitem{aly2008real}
M.~Aly, ``Real time detection of lane markers in urban streets,'' in
  \emph{Intelligent Vehicles Symposium, 2008 IEEE}.\hskip 1em plus 0.5em minus
  0.4em\relax IEEE, 2008, pp. 7--12.

\bibitem{zhao2014real}
J.~Zhao, B.~Xie, and X.~Huang, ``Real-time lane departure and front collision
  warning system on an fpga,'' in \emph{High Performance Extreme Computing
  Conference (HPEC), 2014 IEEE}.\hskip 1em plus 0.5em minus 0.4em\relax IEEE,
  2014, pp. 1--5.

\bibitem{beyeler2014vision}
M.~Beyeler, F.~Mirus, and A.~Verl, ``Vision-based robust road lane detection in
  urban environments,'' in \emph{Robotics and Automation (ICRA), 2014 IEEE
  International Conference on}.\hskip 1em plus 0.5em minus 0.4em\relax IEEE,
  2014, pp. 4920--4925.

\bibitem{chen2017end}
Z.~Chen and X.~Huang, ``End-to-end learning for lane keeping of self-driving
  cars,'' in \emph{Intelligent Vehicles Symposium (IV), 2017 IEEE}.\hskip 1em
  plus 0.5em minus 0.4em\relax IEEE, 2017, pp. 1856--1860.

\bibitem{caltagirone2017fast}
L.~Caltagirone, S.~Scheidegger, L.~Svensson, and M.~Wahde, ``Fast lidar-based
  road detection using convolutional neural networks,'' \emph{IEEE Intelligent
  Vehicles Symposium 2017}, 2017.

\bibitem{laddha2016map}
A.~Laddha, M.~K. Kocamaz, L.~E. Navarro-Serment, and M.~Hebert,
  ``Map-supervised road detection,'' in \emph{Intelligent Vehicles Symposium
  (IV), 2016 IEEE}.\hskip 1em plus 0.5em minus 0.4em\relax IEEE, 2016, pp.
  118--123.

\bibitem{oliveira2017efficient}
G.~L. Oliveira, C.~Bollen, W.~Burgard, and T.~Brox, ``Efficient and robust deep
  networks for semantic segmentation,'' \emph{The International Journal of
  Robotics Research}, p. 0278364917710542, 2017.

\bibitem{vitor2014probabilistic}
G.~B. Vitor, A.~C. Victorino, and J.~V. Ferreira, ``A probabilistic
  distribution approach for the classification of urban roads in complex
  environments,'' in \emph{IEEE Workshop on International Conference on
  Robotics and Automation}, 2014.

\bibitem{einecke2014block}
N.~Einecke and J.~Eggert, ``Block-matching stereo with relaxed fronto-parallel
  assumption,'' in \emph{Intelligent Vehicles Symposium Proceedings, 2014
  IEEE}.\hskip 1em plus 0.5em minus 0.4em\relax IEEE, 2014, pp. 700--705.

\bibitem{Road_Lane_Detection_Survey1}
A.~B. Hillel, R.~Lerner, D.~Levi, and G.~Raz, ``Recent progress in road and
  lane detection: a survey,'' \emph{Machine vision and applications}, vol.~25,
  no.~3, pp. 727--745, 2014.

\bibitem{xiao2015crf}
L.~Xiao, B.~Dai, D.~Liu, T.~Hu, and T.~Wu, ``Crf based road detection with
  multi-sensor fusion,'' in \emph{Intelligent Vehicles Symposium (IV), 2015
  IEEE}.\hskip 1em plus 0.5em minus 0.4em\relax IEEE, 2015, pp. 192--198.

\bibitem{xiao2017hybrid}
L.~Xiao, R.~Wang, B.~Dai, Y.~Fang, D.~Liu, and T.~Wu, ``Hybrid conditional
  random field based camera-lidar fusion for road detection,''
  \emph{Information Sciences}, 2017.

\bibitem{ADAS_Survey}
R.~Okuda, Y.~Kajiwara, and K.~Terashima, ``A survey of technical trend of adas
  and autonomous driving,'' in \emph{VLSI Technology, Systems and Application
  (VLSI-TSA), Proceedings of Technical Program-2014 International Symposium
  on}.\hskip 1em plus 0.5em minus 0.4em\relax IEEE, 2014, pp. 1--4.

\bibitem{kong2010general}
H.~Kong, J.-Y. Audibert, and J.~Ponce, ``General road detection from a single
  image,'' \emph{IEEE Transactions on Image Processing}, vol.~19, no.~8, pp.
  2211--2220, 2010.

\bibitem{mendes2016exploiting}
C.~C.~T. Mendes, V.~Fr{\'e}mont, and D.~F. Wolf, ``Exploiting fully
  convolutional neural networks for fast road detection,'' in \emph{Robotics
  and Automation (ICRA), 2016 IEEE International Conference on}.\hskip 1em plus
  0.5em minus 0.4em\relax IEEE, 2016, pp. 3174--3179.

\bibitem{passani2014crf}
M.~Passani, J.~J. Yebes, and L.~M. Bergasa, ``Crf-based semantic labeling in
  miniaturized road scenes,'' in \emph{Intelligent Transportation Systems
  (ITSC), 2014 IEEE 17th International Conference on}.\hskip 1em plus 0.5em
  minus 0.4em\relax IEEE, 2014, pp. 1902--1903.

\bibitem{huang2009finding}
A.~S. Huang, D.~Moore, M.~Antone, E.~Olson, and S.~Teller, ``Finding multiple
  lanes in urban road networks with vision and lidar,'' \emph{Autonomous
  Robots}, vol.~26, no.~2, pp. 103--122, 2009.

\bibitem{huval2015empirical}
B.~Huval, T.~Wang, S.~Tandon, J.~Kiske, W.~Song, J.~Pazhayampallil,
  M.~Andriluka, P.~Rajpurkar, T.~Migimatsu, R.~Cheng-Yue \emph{et~al.}, ``An
  empirical evaluation of deep learning on highway driving,'' \emph{arXiv
  preprint arXiv:1504.01716}, 2015.

\bibitem{romera2017efficient}
E.~Romera, J.~M. Alvarez, L.~M. Bergasa, and R.~Arroyo, ``Efficient convnet for
  real-time semantic segmentation,'' in \emph{Intelligent Vehicles Symposium
  (IV), 2017 IEEE}.\hskip 1em plus 0.5em minus 0.4em\relax IEEE, 2017, pp.
  1789--1794.

\bibitem{wang2013fpga}
W.~Wang and X.~Huang, ``An fpga co-processor for adaptive lane departure
  warning system,'' in \emph{Circuits and Systems (ISCAS), 2013 IEEE
  International Symposium on}.\hskip 1em plus 0.5em minus 0.4em\relax IEEE,
  2013, pp. 1380--1383.

\bibitem{KITTI_Road_Benchmark}
J.~Fritsch, T.~Kuehnl, and A.~Geiger, ``A new performance measure and
  evaluation benchmark for road detection algorithms,'' in \emph{International
  Conference on Intelligent Transportation Systems (ITSC)}, 2013.

\bibitem{chen2017lidar}
L.~Chen, J.~Yang, and H.~Kong, ``Lidar-histogram for fast road and obstacle
  detection,'' in \emph{Robotics and Automation (ICRA), 2017 IEEE International
  Conference on}.\hskip 1em plus 0.5em minus 0.4em\relax IEEE, 2017, pp.
  1343--1348.

\bibitem{shinzato2014road}
P.~Y. Shinzato, D.~F. Wolf, and C.~Stiller, ``Road terrain detection: Avoiding
  common obstacle detection assumptions using sensor fusion,'' in
  \emph{Intelligent Vehicles Symposium Proceedings, 2014 IEEE}.\hskip 1em plus
  0.5em minus 0.4em\relax IEEE, 2014, pp. 687--692.

\end{thebibliography}

\end{document}